\documentclass[5p]{elsarticle}

\usepackage{lineno}
\modulolinenumbers[5]

\usepackage[T1]{fontenc}
\usepackage{lmodern}

\usepackage{microtype}

\usepackage{flushend}

\usepackage{graphicx}
\usepackage{amsmath,amssymb}
\usepackage{natbib}
\usepackage{url}
\usepackage{hyperref}
\hypersetup{unicode=false,pdfencoding=auto}
\usepackage{booktabs}
\usepackage{array}
\usepackage{xcolor}

\journal{Progress in Retinal and Eye Research}
\begin{document}

\begin{frontmatter}

%% --- Title ---
\title{Myopia Prevention and Control 3.0: Artificial Intelligence--Driven Risk Stratification, Proactive Monitoring, and Personalized Intervention}

%% --- Authors ---
\author[1]{Tieniu Wang\corref{cor1}}
\ead{nltech@163.com}
\author[1,2]{Cangzhu Huang}
\author[2]{Qianhui Li}
\address[1]{Shanghai Nile Intelligent Technology Co., Ltd.}
\address[2]{Beijing Tanyuan Academy of Intelligent Sensing}
\cortext[cor1]{Corresponding author.}

%% --- Abstract ---
\begin{abstract}
The first quarter of the 21st century has witnessed the synchronous 
maturation of artificial intelligence (AI), digital sensing technologies, 
and ubiquitous computing---creating an unprecedented opportunity to 
transform myopia prevention and control from a reactive, population-based 
model into a proactive, precision-driven one. Despite mounting evidence 
that the global myopia burden will reach half the world's population by 
2050, conventional approaches---school-based vision screening (Phase 1.0) 
and evidence-based risk factor management (Phase 2.0)---have proven 
insufficient to alter this trajectory. These limitations arise from 
fundamental characteristics: reactive timing that detects myopia only 
after onset, population-level orientation that applies uniform 
recommendations to heterogeneous individuals, and the absence of a 
continuous feedback loop between risk assessment and intervention. We 
review the emergence of a third paradigm---Myopia Prevention and Control 
3.0---defined by the integration of AI across three interconnected 
domains that form a continuous, closed-loop pipeline. Stage 1 employs 
AI-driven risk stratification to predict individual-level risk of myopia 
onset and progression through machine and deep learning models trained on 
multimodal data. Stage 2 leverages AI-enabled proactive monitoring to 
continuously track behavioural, environmental, and ocular parameters via 
wearable sensors, smartphone applications, and school-based screening 
networks, detecting early signals of change. Stage 3 delivers AI-powered 
personalised intervention that selects, titrates, and dynamically adjusts 
treatment strategies through a closed-loop feedback mechanism. We 
critically evaluate the evidence across each stage, discuss cross-cutting 
challenges in data quality, model validation, ethical governance, and 
health equity, and outline future directions including multimodal 
foundation models, digital twins, and causal machine learning. By framing 
the 3.0 paradigm, we aim to provide a blueprint for researchers, 
clinicians, and policymakers seeking to transform myopia prevention into 
a proactive, precise, and continuously adaptive ecosystem.
\end{abstract}

\begin{keyword}
Myopia prevention and control \sep artificial intelligence \sep 
risk stratification \sep proactive monitoring \sep 
personalized intervention \sep machine learning \sep 
digital health \sep precision public health
\end{keyword}

\end{frontmatter}

%% ====================================================================
%% 1. INTRODUCTION
%% ====================================================================
\section{Introduction}
\label{sec:introduction}

The first quarter of the 21st century has been marked by two 
convergent trends that have set the stage for a fundamental 
re-imagination of myopia prevention and control. On one hand, the 
global myopia epidemic has reached unprecedented scale: the landmark 
study by Holden et al. \cite{holden2016} projected that by 2050, 
approximately 4.76 billion people---half the world's population---will 
be myopic, with 938 million suffering from high myopia. East and 
Southeast Asia bear a disproportionate burden, with school-leaving 
myopia rates exceeding 80--90\% in countries such as Singapore, South 
Korea, and parts of China \citep{morgan2017preer}. The economic 
consequences are staggering: direct healthcare costs and productivity 
losses from uncorrected refractive error and myopic complications are 
estimated in the hundreds of billions of dollars annually 
\citep{baird2020, jonas2017lancet}. On the other hand, the synchronous 
maturation of three technology streams---artificial intelligence (AI), 
digital sensing, and ubiquitous computing---has created, for the first 
time, the technological infrastructure needed to address the myopia 
epidemic with precision, at scale, and in real time.

The prevention and control of myopia have evolved through two major 
phases, yet both have been constrained by their fundamental design. 
Phase 1.0---school-based vision screening---has been the cornerstone of 
public health approaches to myopia for nearly a century. Its logic is 
straightforward: detect reduced visual acuity at fixed intervals and 
refer children for optical correction \citep{morgan2017preer}. This 
approach is, however, inherently reactive: it identifies children who 
are already myopic, cannot predict future onset, and offers no guidance 
for targeted prevention. Phase 2.0---evidence-based risk factor 
management---emerged from two decades of epidemiological research 
demonstrating the protective effects of outdoor time and the risks 
associated with near-work intensity \citep{xiong2017, wu2018, 
gajjar2021}. It introduced clinical interventions with proven efficacy: 
low-dose atropine, orthokeratology, and multifocal contact lenses 
\citep{sankaridurg2023imi, bullimore2025imi}, synthesised through the 
International Myopia Institute (IMI) white papers 
\citep{morgan2021imi, jones2021imi}. Yet Phase 2.0 remains fundamentally 
a population-level paradigm: children receive uniform recommendations 
despite vast individual differences in genetic susceptibility, 
environmental exposure, and treatment response \citep{bullimore2025imi}. 
Adherence to behavioural recommendations is poor, clinical outcomes vary 
widely, and there is no mechanism for continuously adjusting interventions 
based on individual response \citep{wong2020}.

Over the past five years, a convergence of advances in machine learning, 
digital sensing, and ubiquitous computing has opened a fundamentally new 
path forward. In ophthalmology, AI systems have demonstrated diagnostic 
performance comparable to or exceeding that of human experts across 
multiple diseases, including diabetic retinopathy \citep{ting2019preer, 
esteva2021}, age-related macular degeneration \citep{schmidterfurth2018}, 
and glaucoma \citep{li2020preer}. Critically, AI has recently been 
applied directly to the myopia challenge with promising results: deep 
learning models now predict myopia onset \citep{qi2024, gao2025, 
ron2025} and progression to high myopia \citep{foo2023} from fundus 
images and clinical data; machine learning algorithms integrate multi-
modal risk factors for individual-level prediction \citep{zhang2025fundus, 
xi2025, chen2025cyclo}; and digital monitoring systems capture 
behavioural and environmental data at a granularity previously 
unattainable \citep{nurozler2025, abbasi2025}. These developments have 
created the technological prerequisites for a new phase that differs not 
in degree but in kind from its predecessors.

This convergence signals the emergence of a third paradigm---Myopia 
Prevention and Control 3.0---defined by the integration of AI across 
three interconnected domains that together form a continuous, closed-
loop pipeline. The pipeline begins with AI-driven risk stratification 
(Section 3), which answers the question ``Who needs care, at what 
level of intensity?'' by integrating multimodal data through predictive 
models. It proceeds to AI-enabled proactive monitoring (Section 4), 
which answers ``How are risk profiles and physiological parameters 
changing over time?'' by continuously collecting behavioural, 
environmental, and ocular data. It culminates in AI-powered 
personalised intervention (Section 5), which answers ``What should be 
done for this individual, and how should treatment be adjusted?'' by 
selecting and dynamically adapting management strategies. Crucially, 
these three domains are not independent pillars but successive stages of 
a single process, connected by feed-forward and feed-back relationships 
that create---for the first time in myopia prevention---a true closed-
loop system.

In this review, we provide a comprehensive synthesis of the emerging 
evidence for the 3.0 paradigm, organised around this closed-loop 
framework. Section 2 traces the historical evolution from 1.0 to 3.0, 
highlighting the conceptual and technological discontinuities that 
separate them. Sections 3, 4, and 5 examine the evidence for risk 
stratification, proactive monitoring, and personalised intervention, 
respectively, with explicit attention to how each stage connects to the 
others. Section 6 discusses cross-cutting challenges and future 
directions. Section 7 concludes with a synthesis of the 3.0 paradigm 
and a call for interdisciplinary collaboration to realise its potential.

%% --- Figure 1: Evolution timeline ---
%% Gemini-generated (v5): Three-panel timeline 1.0 (Passive Correction) -> 2.0 (Active Intervention) -> 3.0 (Precision Management)
%% Dashed border on 3.0 panel with "Proposed Framework" tag.
\begin{figure*}[hbtp!]
\centering
\includegraphics[width=\textwidth]{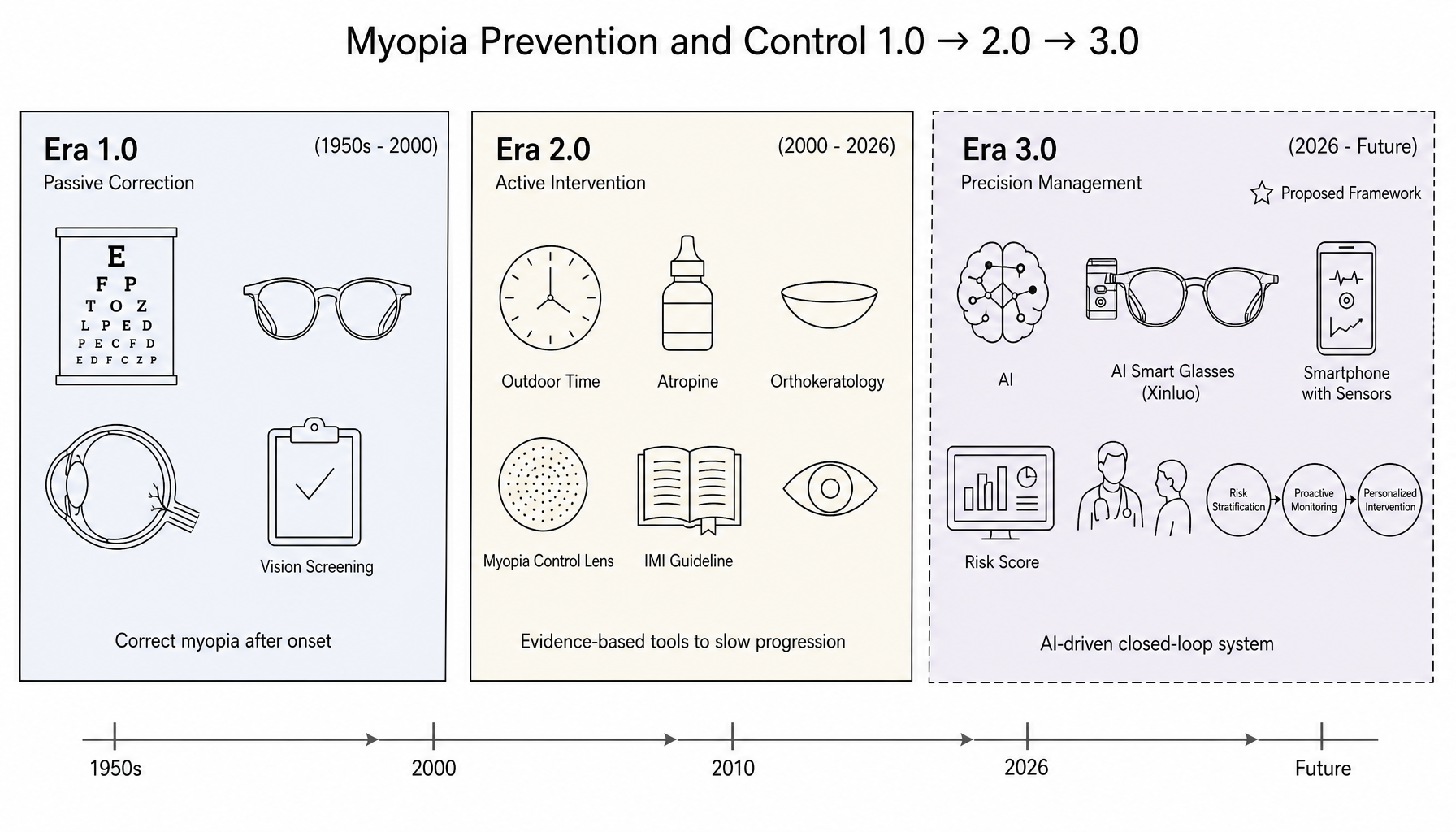}
\caption{%
  Evolution of myopia prevention and control from 1.0 to 3.0.
  Era~1.0 (Passive Correction, 1950s--2000) relied on school vision screening
  and optical correction. Era~2.0 (Active Intervention, 2000--2026) introduced
  evidence-based strategies (atropine, orthokeratology, outdoor time). Era~3.0
  (AI-Driven Precision Management, proposed) integrates risk stratification,
  wearable monitoring, and personalised intervention in a closed-loop framework.
  The 3.0 panel is shown with a dashed border to indicate its proposed nature.}
\label{fig:evolution}
\end{figure*}

%% ====================================================================
%% 2. THE EVOLUTION OF MYOPIA PREVENTION: FROM 1.0 TO 3.0
%% ====================================================================
\section{The Evolution of Myopia Prevention and Control: From 1.0 to 3.0}
\label{sec:evolution}

To understand the transformative potential of the 3.0 paradigm, it is 
instructive to trace the historical arc of myopia prevention---from 
reactive screening (1.0), through evidence-based risk management (2.0), 
to the emerging AI-driven precision ecosystem (3.0). Table 1 summarizes 
the key characteristics of each era.

\begin{table*}[htbp!]
\centering
\caption{Comparison of myopia prevention and control paradigms: 1.0, 2.0, and 3.0.}
\label{tab:evolution}
\footnotesize
\begin{tabular}{p{2.0cm} p{3.5cm} p{3.5cm} p{3.5cm}}
\toprule
\textbf{Dimension} & \textbf{1.0: Screening Era} 
& \textbf{2.0: Intervention Era} 
& \textbf{3.0: AI-Driven Era} \\
\midrule
Core approach & Vision screening + glasses & Evidence-based risk management & AI-driven precision pipeline \\
Target & All school children & At-risk groups & Every individual \\
Timing & Reactive (after onset) & Early (pre-myopia) & Proactive (predictive) \\
Data source & Visual acuity & Research cohort data & Multi-modal, continuous \\
Decision basis & Fixed thresholds & Group-level evidence & Individual-level prediction \\
Intervention & Optical correction & Standardized protocols & Personalized, adaptive \\
Feedback loop & None & Disconnected & Closed-loop (5$\to$4$\to$5) \\
Key technology & Snellen chart & Clinical trials & ML/DL, wearables, IoT \\
\bottomrule
\end{tabular}
\end{table*}

\subsection{Era 1.0: Population-Based Vision Screening (20th Century)}
\label{sec:1.0}

The origins of organised myopia prevention trace to school-based vision 
screening programmes, which became widespread across East Asia, Europe, 
and North America throughout the twentieth century. These programmes 
typically involve periodic visual acuity (VA) assessment using Snellen 
charts, with children who fail a fixed threshold (e.g., VA $\leq$ 6/12 
or 20/40) referred for cycloplegic refraction and optical correction 
\citep{morgan2017preer}. The underlying logic is straightforward: 
identify children who already have clinically significant refractive 
error and provide them with spectacles.

While this model has achieved notable successes---increasing the 
proportion of myopic children who receive correction and, in some 
settings, improving educational outcomes---its fundamental limitations 
have become increasingly apparent. First, vision screening is inherently 
reactive: it can only detect myopia after it has already developed, and 
often after significant progression has occurred. Second, standard VA 
tests have limited sensitivity for detecting early or low-grade myopia, 
particularly in young children who may not report visual symptoms 
\citep{jones2021imi}. Third, screening intervals are typically annual or 
biennial, creating windows of undetected progression during which axial 
length can increase substantially. Finally, the screening approach 
treats all children uniformly, offering no mechanism for identifying 
those at highest risk before the onset of myopia. In essence, Era 1.0 
addresses the question ``Who is already myopic?'' rather than the more 
clinically relevant question ``Who will become myopic and needs 
preventive intervention?''

\subsection{Era 2.0: Evidence-Based Risk Factor Management (2000-2026)}
\label{sec:2.0}

The turn of the millennium marked a shift from detection alone toward 
understanding and modifying the risk factors for myopia onset and 
progression. Two decades of epidemiological research established outdoor 
time as the most consistently modifiable protective factor. The protective 
effect of time outdoors---mediated by bright light stimulating retinal 
dopamine release, which inhibits axial elongation---was demonstrated across 
multiple independent cohorts and clinical trials 
\citep{xiong2017, wu2018, read2019}. Concurrently, near-work intensity 
and duration were identified as independent risk factors, although their 
effect sizes are smaller than initially assumed 
\citep{gajjar2021, biswas2024}.

The establishment of the International Myopia Institute (IMI) in 2015 
consolidated Phase 2.0 by producing authoritative white papers on myopia 
definitions 
\citep{morgan2021imi}, interventions 
\citep{sankaridurg2023imi, bullimore2025imi}, and risk factors 
\citep{jones2021imi}. These evidence syntheses translated research into 
practice guidelines that formed the basis for national myopia prevention 
programmes in Singapore, China, Taiwan, and other countries 
\citep{morgan2017preer, jonas2017lancet}. Clinical interventions---low-dose 
atropine (0.01\%), orthokeratology (OK), multifocal contact lenses (MFCLs), 
and peripheral defocus spectacles---demonstrated efficacy in slowing 
progression, with treatment effects varying from 30\% to 70\% reduction 
in axial elongation 
\citep{sankaridurg2023imi, bullimore2025imi}.

Despite these advances, Phase 2.0 has encountered fundamental barriers. 
First, the evidence-to-practice gap remains wide: even in countries with 
national programmes, adherence to outdoor-time recommendations is poor 
\citep{wong2020}. Second, intervention effects are highly heterogeneous---
some children respond well to low-dose atropine while others show minimal 
response, yet the 2.0 paradigm lacks tools to predict individual response 
\citep{bullimore2025imi}. Third, monitoring of risk factors and treatment 
effects remains episodic, relying on periodic clinic visits that capture 
snapshots rather than trajectories. These limitations created a growing 
recognition that population-level recommendations, while necessary, are 
insufficient for achieving individual-level prevention.

\subsection{Era 3.0: AI-Driven Precision Myopia Management (2026--Future)}
\label{sec:3.0}

Phase 3.0 represents a paradigm shift from population-level risk management 
to individual-level precision care, enabled by the confluence of three 
technological currents: (1) advances in machine learning and deep learning 
that can model complex, non-linear relationships in multi-modal data; 
(2) the proliferation of digital sensing devices including smartphones, 
wearables, and smart environments that generate continuous behavioural 
and physiological data streams; and (3) the maturation of cloud computing 
and Internet-of-Things (IoT) infrastructure that can support real-time, 
scalable data processing 
\citep{li2020preer, ting2019preer}.

What distinguishes 3.0 from its predecessors is not merely the addition 
of AI tools but the fundamental re-architecture of the prevention 
workflow. In 1.0, decision-making was provider-driven and episodic: a 
school nurse or optometrist assessed vision at fixed intervals. In 2.0, 
decision-making became evidence-informed but still population-centred: 
clinical practice guidelines prescribed the same outdoor-time target or 
atropine concentration for broad age groups. In 3.0, decision-making is 
data-driven, continuous, and individualised: the system learns from each 
child's data to predict their trajectory, monitor their progress, and 
adapt their care in real time (see Table 1).

\begin{figure*}[htbp!]
\centering
\includegraphics[width=\textwidth]{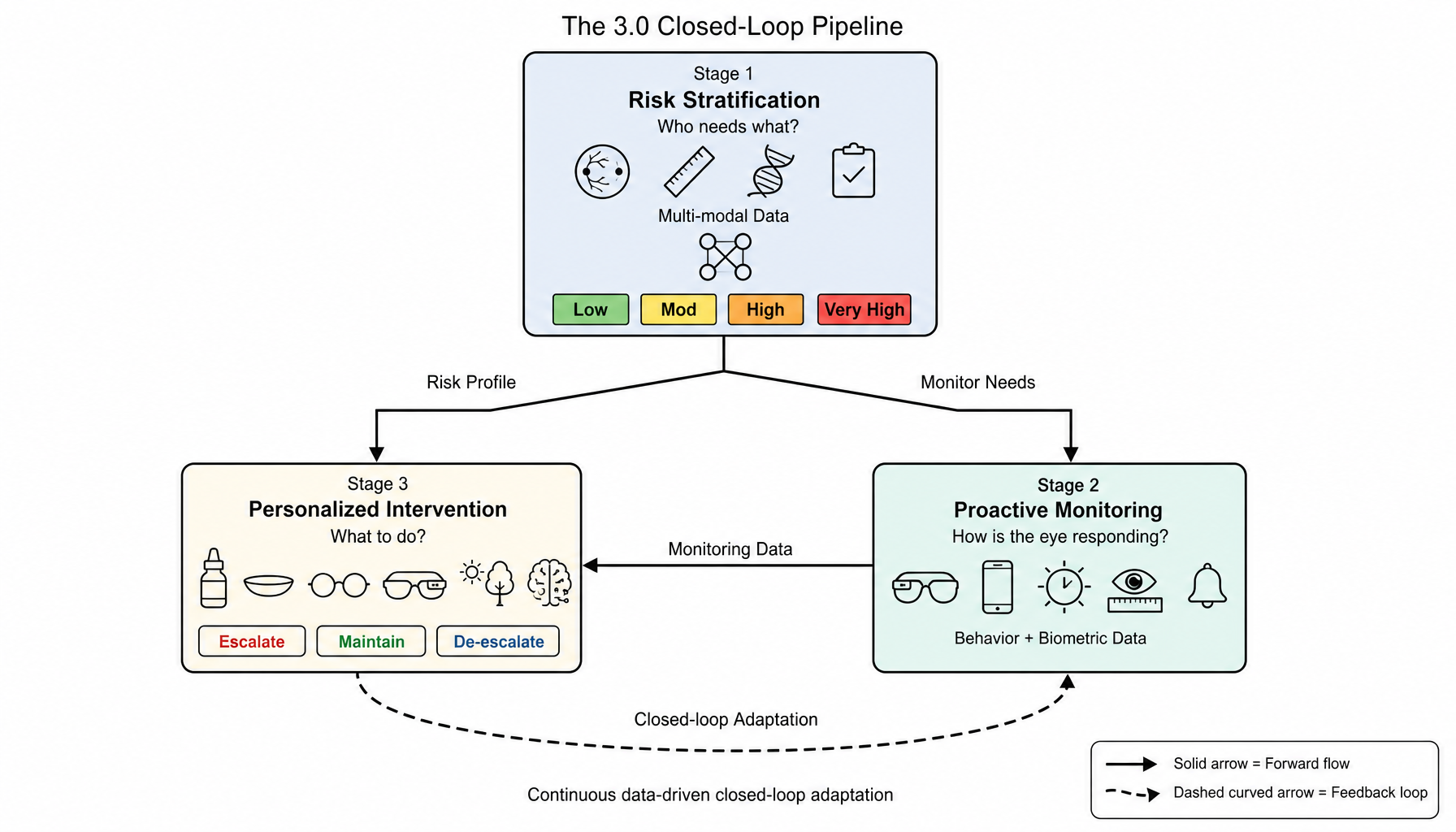}
\caption{%
  The 3.0 closed-loop pipeline. Risk stratification generates
  an individualised risk profile that determines monitoring intensity. Monitoring
  alerts trigger intervention adjustments, and outcomes feed back into
  monitoring, forming a continuous adaptive loop.}
\label{fig:pipeline}
\end{figure*}

The 3.0 paradigm is defined by the closed-loop pipeline that constitutes 
the central framework of this review (Fig.~\ref{fig:pipeline}). The pipeline begins with 
AI-driven risk stratification (Section 3) to answer ``Who needs what 
level of preventive care?''. It proceeds to AI-enabled proactive 
monitoring (Section 4) to answer ``How is this individual responding to 
current conditions and interventions?''. It culminates in personalised 
intervention (Section 5) to answer ``What specific action is optimal for 
this person now?''. Critically, the output of Section 5 feeds back into 
Section 4, creating a dynamic closed loop in which intervention effects 
are continuously assessed and adjustments are made in response. It is 
this feedback mechanism---entirely absent from both 1.0 and 2.0---that 
constitutes the central innovation of the 3.0 paradigm.

%% --- Figure 2: The 3.0 closed-loop pipeline (core conceptual framework) ---
%% Required: A three-stage cycle (Risk Stratification -> Proactive Monitoring -> Personalized Intervention -> back),
%% with clear feed-forward and feedback arrows. This is the most important figure in the paper.

%% ====================================================================
%% 3. AI-DRIVEN RISK STRATIFICATION
%% ====================================================================
\section{AI-Driven Risk Stratification}
\label{sec:risk-stratification}

\textbf{Role in the 3.0 pipeline:} Risk stratification is the entry 
point of the closed-loop system. Its output---a dynamic risk label 
(low, moderate, high, or very high) for each individual---determines 
the intensity of monitoring (Section 4) and the aggressiveness of 
intervention (Section 5). Without accurate stratification, downstream 
efforts become unfocused.

\subsection{Why Risk Stratification Matters: The Entry Point}
\label{sec:3.1}

Risk stratification is the gateway to the 3.0 pipeline. Its function is 
to answer a question that neither 1.0 nor 2.0 could address with 
adequate precision: among the billions of children worldwide, which 
individuals are at sufficiently high risk of developing or progressing 
toward sight-threatening myopia that they warrant intensive monitoring 
and early intervention?

Traditional risk scores for myopia have relied on a small set of 
established factors: parental myopia, age of onset, baseline spherical 
equivalent, and time outdoors 
\citep{morgan2021imi, jones2021imi}. While these factors are 
epidemiologically robust at the population level, their discriminative 
power at the individual level is modest. A recent systematic review of 
myopia prediction models reported C-statistics ranging from 0.65 to 0.85, 
with most models achieving only moderate discrimination 
\citep{han2021risk}. The central challenge is that myopia arises from 
complex interactions between genetic susceptibility, environmental 
exposure, and behavioural patterns---interactions that linear risk scores 
cannot capture.

AI-based risk stratification addresses this challenge in several ways. 
First, machine learning models can incorporate high-dimensional data 
from multiple modalities, identifying non-linear interactions that 
traditional regression approaches would miss. Second, deep learning 
models can extract predictive features directly from raw data such as 
fundus images, bypassing the need for manually engineered features. 
Third, AI models can be updated dynamically as new data accumulate, 
moving from a static baseline prediction to a continuously refined risk 
profile. The remainder of this section reviews the evidence for AI-driven 
risk stratification organised by prediction target (onset vs. progression) 
and data modality.

\subsection{Machine Learning Models for Onset and Progression Prediction}
\label{sec:3.2}

The literature on ML-based myopia risk prediction has grown rapidly since 
2023, spanning diverse model architectures, data sources, and prediction 
time horizons. The studies reviewed here are summarised in Table 2.

Several recent studies have focused on predicting myopia onset in 
pre-myopic children. Qi et al. (2024) developed a deep learning model 
integrating fundus photographs and clinical data to predict both the 
likelihood of myopia onset within three years and the expected refractive 
error trajectory, achieving an area under the curve (AUC) of 0.88 in 
external validation \citep{qi2024}. Gao et al. (2025) applied machine 
learning algorithms to a longitudinal cohort of 2,375 pre-myopic children 
in China, demonstrating that ensemble methods combining gradient-boosted 
trees with clinical variables outperformed conventional logistic 
regression (AUC 0.84 vs. 0.70) \citep{gao2025}. Ron et al. (2025) 
leveraged routine eye examination data from a large Israeli cohort to 
develop a ML-based myopia prediction tool, reporting comparable 
performance to more resource-intensive models, highlighting the potential 
of using routinely collected clinical data \citep{ron2025}.

Beyond onset, a growing number of studies target prediction of rapid 
progression or progression to high myopia---a clinically more consequential 
endpoint. Foo et al. (2023) trained a deep learning system on colour 
fundus photographs to forecast five-year risk of high myopia (spherical 
equivalent $\leq$-5.00 D) in Singaporean children, achieving an AUC of 
0.87 and identifying fundus features such as tessellation and peripapillary 
features as strong predictors \citep{foo2023}. Li et al. (2022) developed 
and validated a ML model using data from the Anyang Childhood Eye Study, 
finding that axial length, baseline refraction, and parental myopia were 
the most important predictors with a random forest model achieving an AUC 
of 0.83 for three-year progression \citep{li2022ml}. Guan et al. (2023) 
demonstrated the feasibility of deploying ML-based progression prediction 
using real-world school screening data, achieving practical accuracy 
without the need for specialised research-grade measurements \citep{guan2023}.

%% --- Figure 3 placeholder: Risk stratification models overview ---
%% Please insert: A visual summary of key ML/DL models for myopia risk prediction,
%%   comparing model type, data modality, prediction target, and performance (AUC).
%%   Could be presented as a figure or supplement to Table 2.
\begin{figure*}[htbp!]
\centering
\includegraphics[width=\textwidth]{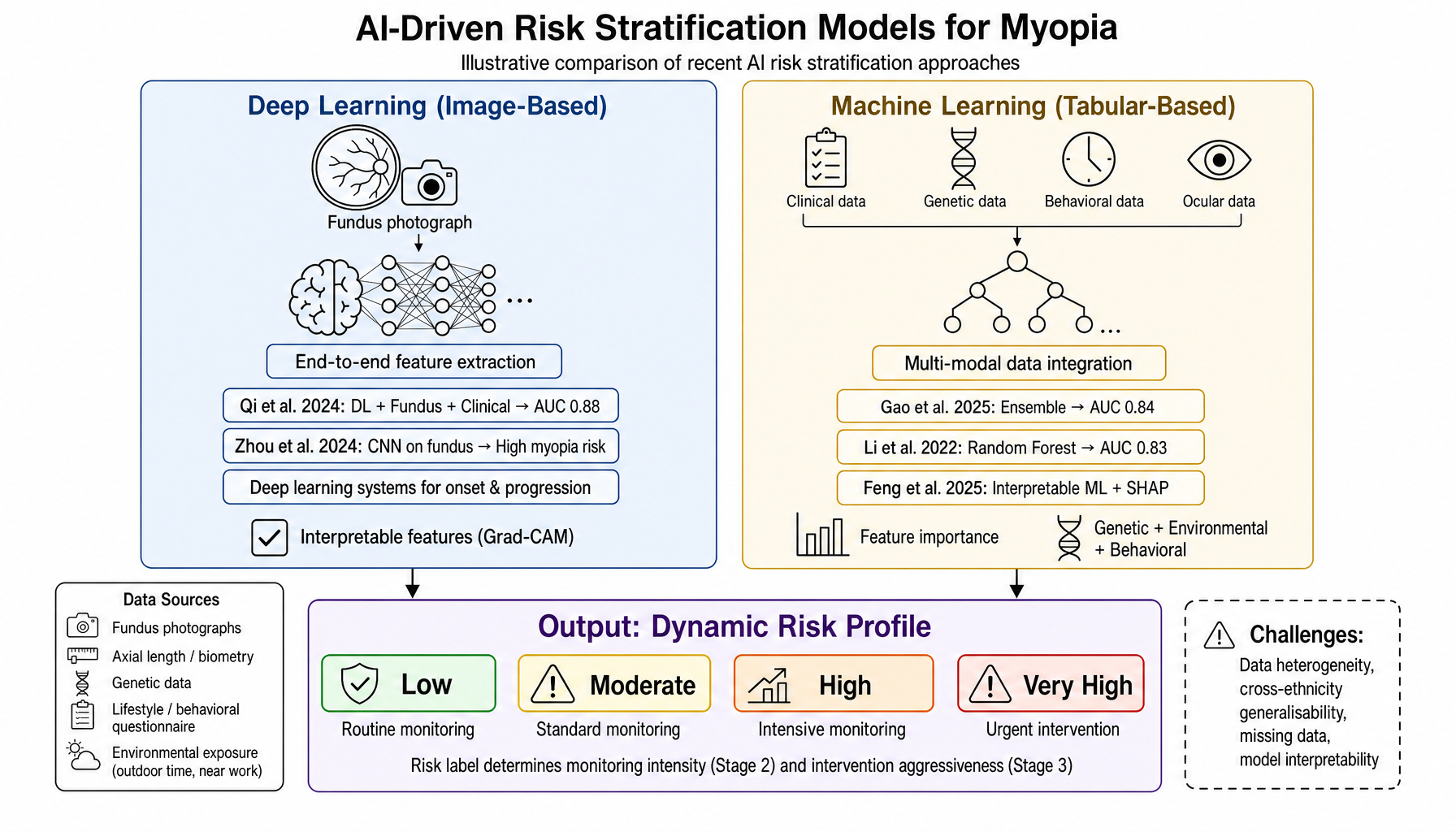}
\caption{%
  Comparison of AI risk stratification approaches. Deep learning models
  predict onset and progression from fundus images; machine learning models
  integrate clinical, environmental, and behavioural variables for
  individualised risk profiling (see Section 3.2).}
\label{fig:risk-models}
\end{figure*}

Feng et al. (2025) specifically addressed the need for interpretability in 
clinical deployment, developing an interpretable ML model from school 
screening data and reporting that SHAP value analysis identified near-work 
duration, parental myopia history, and baseline axial length as the top 
three predictors, consistent with established epidemiological knowledge 
\citep{feng2025}. These findings collectively demonstrate that AI models 
cannot only achieve clinically useful predictive accuracy but can also 
align with and extend our understanding of myopia pathophysiology.

\subsection{Multimodal Integration: Imaging, Environment, and Genetics}
\label{sec:3.3}

A key advantage of AI-based risk stratification over traditional approaches 
is its capacity to integrate heterogeneous data types into a unified 
predictive framework. The strongest evidence for multimodal integration 
comes from studies combining fundus imaging with clinical and behavioural 
variables. Xi et al. (2025) demonstrated that a multi-factorial ML model 
incorporating fundus features, parental history, and lifestyle variables 
significantly outperformed any single-modality model in predicting 
three-year myopia incidence \citep{xi2025}. Similarly, Zhang et al. 
(2025) developed a ``fundus imageomics'' approach that extracted 
quantitative imaging features from colour fundus photographs and combined 
them with clinical risk factors, achieving an AUC of 0.91 for predicting 
high myopia development \citep{zhang2025fundus}.

Chen et al. (2025) focused specifically on the role of cycloplegic 
refractive error components, demonstrating that ML models incorporating 
both anterior segment biometrics and fundus features provided more 
accurate progression predictions than models based on spherical equivalent 
alone \citep{chen2025cyclo}. Liao et al. (2025) evaluated a comprehensive 
set of youth-specific predictors---including near-work patterns, outdoor 
activity duration, sleep habits, and digital device usage---in a machine 
learning framework and found that digital device usage patterns contributed 
independent predictive value beyond traditional near-work measures \citep{liao2025}.

Despite these promising results, multimodal integration faces several 
challenges. Data heterogeneity---combining image, tabular, and time-series 
data---requires specialised fusion architectures. Missing data are common 
in real-world settings, particularly for questionnaire-derived behavioural 
variables. Feature selection in high-dimensional multimodal spaces risks 
overfitting without rigorous external validation. Addressing these 
challenges will require standardised data collection protocols, robust 
missing-data handling methods, and prospective validation studies across 
diverse populations.

\subsection{Dynamic Updating: From Static Screenshot to Living Risk Profile}
\label{sec:3.4}

A fundamental limitation of current risk stratification research is its 
reliance on single-timepoint prediction. Most studies use baseline data 
to predict outcomes years into the future, implicitly assuming that risk 
factors remain static---an assumption that contradicts the dynamic nature 
of myopia development, where behaviour, environment, and physiology 
change continuously.

The vision for 3.0 is that of a ``living risk profile'' that updates 
continuously as new data from monitoring (Section 4) become available. 
This would allow the model to detect risk-increasing behavioural shifts 
(e.g., a sudden increase in near-work during exam periods) and adjust the 
risk label accordingly, potentially triggering earlier or more intensive 
monitoring. While fully dynamic updating systems have not yet been 
demonstrated in the myopia literature, the technical building blocks---
Bayesian updating, online learning, and recurrent neural networks for 
time-series prediction---are well-established in other medical domains. 
Bridging from static to dynamic risk stratification represents one of the 
most important directions for future research, and directly connects risk 
stratification (Section 3) with proactive monitoring (Section 4) in the 
3.0 pipeline.

%% ====================================================================
%% 4. PROACTIVE MONITORING POWERED BY ARTIFICIAL INTELLIGENCE
%% ====================================================================
\section{Proactive Monitoring Powered by Artificial Intelligence}
\label{sec:monitoring}

\textbf{Role in the 3.0 pipeline:} Monitoring is the bridge between risk 
stratification and intervention. It receives risk-stratified individuals 
from Section 3 and continuously collects behavioral, environmental, and 
ophthalmic data to detect deviations from expected trajectories. When 
progression signals are detected, the system triggers adjustments in 
Section 5. Conversely, after an intervention is modified (Section 5), 
monitoring evaluates its effectiveness, closing the loop.

\subsection{Defining the Digital Phenotype: What to Monitor}
\label{sec:4.1}

Central to the monitoring stage is the concept of the digital 
phenotype---the moment-by-moment quantification of an individual's 
behaviour, environment, and physiology through digital sensors. For 
myopia prevention, the key components of the digital phenotype include: 
(1) light exposure (intensity, spectral composition, and circadian 
pattern), (2) near-work activity (duration, distance, and breaks), 
(3) physical activity and outdoor time, and (4) ocular biometric 
parameters when self-monitoring devices are available.

The rationale for monitoring these dimensions is grounded in the 
established pathophysiology of myopia. Bright outdoor light (typically 
$>$ 1000 lux) stimulates retinal dopamine release, which inhibits axial 
elongation through a signalling cascade involving the retina, retinal 
pigment epithelium, and choroid \citep{read2019, ashby2025}. Near-work 
induces accommodative demand and may create peripheral hyperopic defocus, 
a putative stimulus for axial elongation \citep{gajjar2021}. Continuous 
monitoring of these parameters enables quantification of an individual's 
``exposome''---the totality of their environmental exposures over time---
allowing the system to identify windows of elevated risk that would be 
invisible to periodic questionnaires \citep{biswas2024}.

\subsection{Technologies for Continuous Monitoring: Wearables and Smartphones}
\label{sec:4.2}

The technological infrastructure for continuous monitoring has matured 
substantially over the past five years. The most widely studied approach 
uses wrist-worn actigraphy devices (e.g., Actiwatch, Fitbit) to 
discriminate indoor from outdoor time based on ambient light intensity. 
Nurozler Tabakci et al. (2025) demonstrated that consumer-grade 
actigraphy devices can objectively classify time outdoors with over 85\% 
accuracy in children, providing an alternative to subjective parental 
questionnaires that are notoriously unreliable \citep{nurozler2025}. 
Abbasi et al. (2025) extended this approach by developing ML classifiers 
trained on Actiwatch data to categorise indoor vs. outdoor light exposure 
patterns, achieving sensitivity and specificity exceeding 90\% \citep{abbasi2025}. 
Jiang et al. (2025) further investigated the spectral composition of outdoor 
light exposure, finding that both intensity and duration of exposure to 
short-wavelength-enriched light were independently associated with slower 
axial elongation \citep{jiang2025}.

Smartphone-based approaches offer complementary capabilities. The ubiquity 
of smartphones makes them a scalable platform for monitoring near-work 
behaviours. Su et al. (2025) developed a smartphone-based screening 
platform that uses in-built sensors to estimate reading distance, ambient 
illumination, and duration of near-work sessions, providing a low-cost, 
scalable tool for large-scale monitoring \citep{su2025}. Computer vision 
approaches using the front-facing camera can estimate gaze distance and 
angle, while app-based diaries can capture near-work type (digital vs. 
print) and break frequency. The combination of wearable and smartphone 
data creates a multi-sensor monitoring system that captures both the 
environmental context (light, location) and the behavioural response 
(near-work, posture) relevant to myopia risk.

\subsection{AI Analysis of Monitoring Data: Pattern Recognition and Alerting}
\label{sec:4.3}

Raw sensor data, no matter how comprehensive, is of limited clinical value 
without intelligent analysis to extract meaningful patterns and generate 
actionable alerts. AI plays this interpretive role at several levels. At 
the first level, classification algorithms label sensor readings (e.g., 
``outdoor play'' vs. ``indoor near-work'' vs. ``indoor distance viewing'') 
with high temporal resolution. At the second level, pattern recognition 
algorithms identify behavioural phenotypes---clusters of children who 
share similar activity profiles---enabling the system to detect when an 
individual deviates from their typical pattern in a risk-increasing 
direction. For example, an increase in nocturnal light exposure coinciding 
with examination periods may signal both reduced outdoor time and 
disrupted circadian rhythms, each contributing to myopia risk 
\citep{li2023digital}.

At the third level, alerting algorithms integrate monitoring data with the 
risk profile from Section 3 to determine when a change in intervention is 
warranted. These algorithms must balance sensitivity (detecting clinically 
meaningful progression early) against specificity (avoiding unnecessary 
alerts that contribute to alert fatigue). The threshold for triggering an 
alert may be calibrated to the individual's risk stratum: a high-risk child 
may warrant an alert after a smaller deviation than a low-risk child would. This 
risk-calibrated alerting represents the direct feed-forward connection from 
Section 3 to Section 4 in the 3.0 pipeline.

\subsection{School-Based AI-Enhanced Screening Networks}
\label{sec:4.4}

School-based vision screening---a legacy of the 1.0 era---provides an existing 
infrastructure that can be upgraded for the 3.0 paradigm. Traditional school 
screening suffers from known limitations: variable quality across screening 
sites, subjective interpretation of VA results, and lack of longitudinal 
data continuity between grades \citep{morgan2017preer}. AI-enhanced 
screening addresses these gaps in three ways. First, automated refraction 
and fundus imaging can standardise data collection across screening sites, 
reducing operator-dependent variability. Second, cloud-based platforms 
can aggregate data across schools and districts, creating a longitudinal 
record that tracks each child's trajectory across years \citep{guan2023}. 
Guo et al. (2025) demonstrated that AI-based quantification of fundus 
tessellation from screening fundus photographs can serve as a quantitative 
biomarker for high myopia risk, providing objective risk information from 
a single screening visit \citep{guo2025}.

At the regional level, AI-powered surveillance systems can aggregate 
screening data to monitor population-level trends, detect emerging myopia 
hotspots, and allocate resources efficiently. Jin et al. (2025) proposed 
a digital hierarchical vision health management system that integrates 
school screening, community health records, and hospital-based care into 
a unified decision-support platform---an early prototype of the 
comprehensive monitoring infrastructure required for the 3.0 paradigm 
\citep{jin2025}.

\subsection{From Monitoring to Alerting: Triggering Intervention Adjustments}
\label{sec:4.5}

The ultimate purpose of monitoring is not data collection for its own sake 
but to trigger timely adjustments in intervention when monitoring signals 
indicate that the current strategy is insufficient. The key question is: 
what constitutes an actionable signal? Clinically, the most important 
parameters are axial length increase (typically $>$ 0.2 mm/year in children) 
and myopic shift in spherical equivalent ($>$ 0.50 D/year). However, 
behavioural indicators from continuous monitoring can provide earlier 
signals: sustained reductions in outdoor time below a personalised 
threshold, or increases in near-work beyond an individual's typical 
range, may precede detectable axial elongation by weeks or months.

The transition from monitoring to intervention adjustment (the feed-forward 
from Section 4 to Section 5) and from intervention back to monitoring (the 
feedback loop from Section 5 to Section 4) defines the closed-loop nature 
of the 3.0 paradigm. This integration of monitoring and intervention into 
a single, continuously operating system---rather than separate episodic 
activities---differentiates 3.0 from all previous approaches to myopia 
prevention and control \citep{pan2025}.

%% --- Figure 4 placeholder: Monitoring system architecture ---
%% Please insert: Schematic of the AI-enabled proactive monitoring ecosystem showing:
%%   (1) Wearable sensors (actigraphy, light sensors) on the child
%%   (2) Smartphone-based near-work tracking
%%   (3) School-based screening network with cloud data aggregation
%%   (4) AI analysis layer with pattern recognition and alerting
%%   (5) Feedback to clinician and parents
\begin{figure*}[htbp!]
\centering
\includegraphics[width=\textwidth]{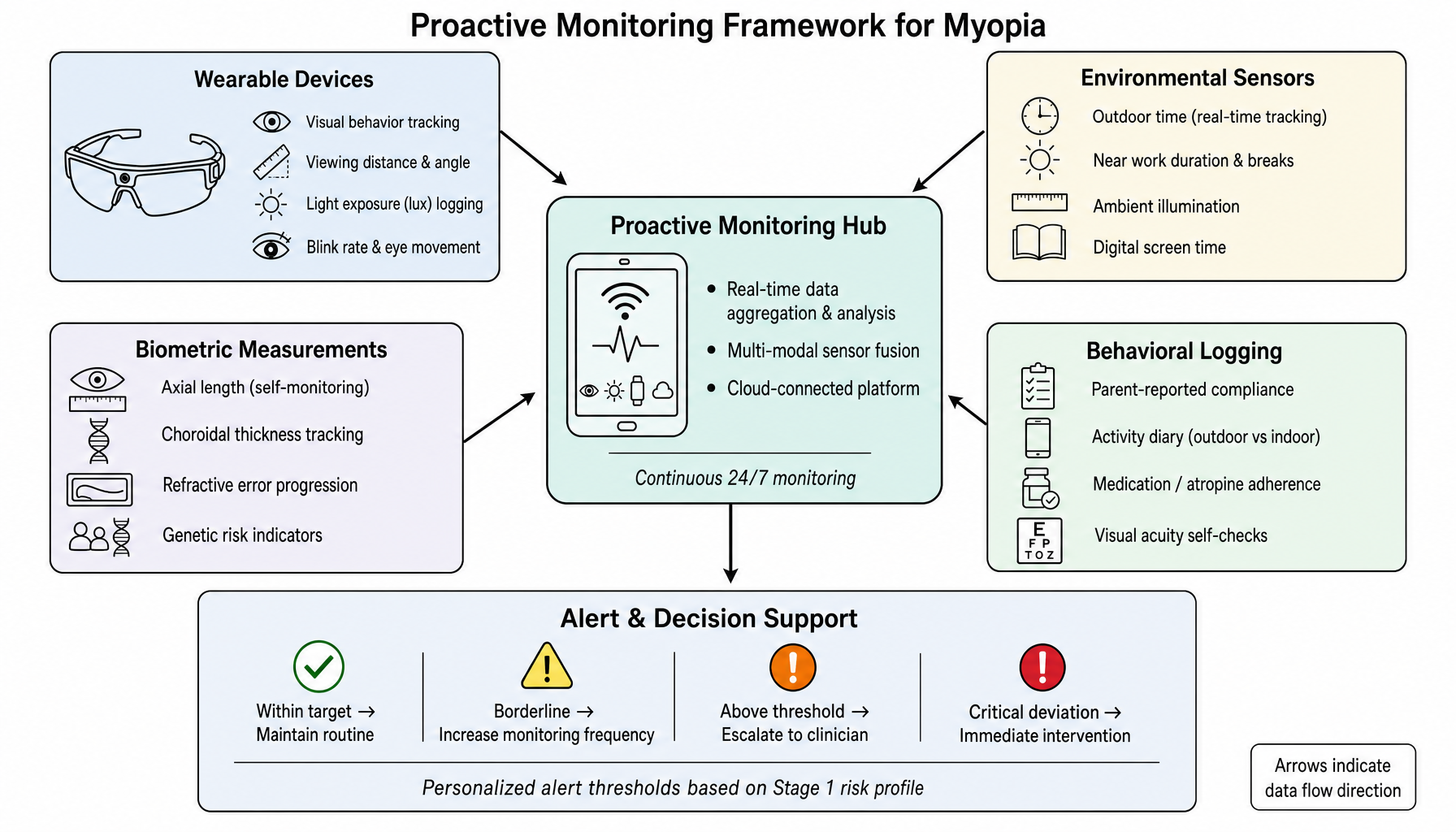}
\caption{%
  Proactive monitoring architecture. Multi-sensor data from wearables,
  smartphones, and school screening are processed through AI pipelines for
  behavioural phenotyping, anomaly detection, and risk-calibrated alert
  generation. Outputs update the risk profile and guide intervention.}
\label{fig:monitoring}
\end{figure*}

%% ====================================================================
%% 5. PERSONALIZED INTERVENTION STRATEGIES
%% ====================================================================
\section{Personalized Intervention Strategies}
\label{sec:intervention}

\textbf{Role in the 3.0 pipeline:} This is the action stage of the 
loop. It receives two inputs: (1) the risk profile from Section 3 
(who needs what level of intervention) and (2) the continuous 
monitoring data from Section 4 (how the individual's condition is 
evolving). Its output is an individualized treatment plan that is 
dynamically adjusted as new monitoring data arrives.

\subsection{Why Personalization Matters: Beyond One-Size-Fits-All}
\label{sec:5.1}

The transition from 2.0 to 3.0 is perhaps most consequential in the 
intervention domain. Phase 2.0 has provided a growing toolbox of 
effective interventions: low-dose atropine, orthokeratology, multifocal 
contact lenses, and peripheral defocus spectacles. However, the standard 
approach has been to select an intervention based on age and baseline 
refraction alone---a one-size-fits-all strategy that produces highly 
variable outcomes. The IMI Intervention Digests report treatment effect 
reductions in axial elongation ranging from 30\% to 70\%, with substantial 
inter-individual variation unexplained by current clinical criteria 
\citep{sankaridurg2023imi, bullimore2025imi}.

The 3.0 approach to intervention personalisation differs in two 
fundamental ways. First, it tailors not only the type of intervention 
but also its intensity to the individual's risk profile from Section 3. 
A child with a low predicted progression risk may receive a behavioural 
intervention alone, while a high-risk child may receive combination 
therapy with pharmacologic and optical components. Second, it adjusts 
the intervention dynamically based on monitoring data from Section 4, 
in a continuous feedback loop rather than at fixed follow-up intervals. 
This section reviews the emerging evidence on AI-assisted treatment 
selection, digital therapeutics, and the closed-loop feedback mechanism 
that connects intervention with monitoring.

\subsection{AI-Assisted Treatment Selection: Predicting Response}
\label{sec:5.2}

A growing body of research demonstrates that ML models can predict 
individual treatment response, enabling clinicians to select the 
intervention most likely to succeed for each patient before initiating 
therapy. Rong et al. (2025) developed a ML model to predict axial length 
response to orthokeratology using baseline corneal topography and 
clinical parameters, achieving a mean absolute prediction error of 
0.11 mm for 12-month axial length change---sufficiently accurate to 
distinguish likely responders from non-responders and potentially avoid 
the cost and inconvenience of a three-month trial period \citep{rong2025}. 
Zhang et al. (2025) applied similar methods to predict OK efficacy, 
identifying flatter corneal curvature and higher baseline myopia as 
predictors of better response \citep{zhang2025ok}.

For pharmacologic interventions, Chen et al. (2025) developed a decision 
tree algorithm to predict response to low-dose atropine (0.01\%) based 
on age, baseline refraction, and early treatment response (3-month axial 
length change), achieving 78\% accuracy in predicting 2-year treatment 
success \citep{chen2025atropine}. These prediction tools have direct 
clinical utility: a child predicted to have poor atropine response could 
be considered for combination therapy or alternative optical interventions 
from the outset, avoiding months of ineffective treatment. Extending this 
logic, future ``treatment recommender systems'' could compare predicted 
responses across multiple intervention options and rank them for each 
individual, analogous to how streaming services recommend content based 
on user profiles.

\subsection{Digital Therapeutics and Behavioural Modification}
\label{sec:5.3}

Digital therapeutics---evidence-based therapeutic interventions driven 
by software---represent a rapidly expanding category of myopia 
interventions that align naturally with the 3.0 paradigm. These include 
virtual reality (VR) and augmented reality (AR) training systems designed 
to enhance accommodative function or impose peripheral myopic defocus. 
Ping et al. (2025) conducted a systematic review of VR interventions for 
myopia, finding preliminary evidence that VR-based visual training can 
improve accommodative amplitude and reduce subjective visual fatigue, 
though longer-term effects on axial elongation remain to be demonstrated 
\citep{ping2025}. Hu et al. (2025) investigated AR visual training and 
reported significant increases in choroidal thickness following 
training sessions, suggesting a potential mechanism for myopia control 
\citep{hugr2025}.

Beyond device-based interventions, AI-powered behavioural modification 
platforms can deliver personalised nudges based on monitoring data. A 
recent cluster-randomised trial by Hu et al. showed that wearable 
eye-use monitoring combined with vibration alerts and behavioural 
feedback significantly reduced myopia onset and progression in children,
providing clinical evidence for the efficacy of this approach 
\citep{hu2025behavioral}. For example, when the monitoring system 
detects that a child's daily outdoor time has fallen below their 
personalised threshold for three consecutive days, the platform can send 
a tailored reminder to parents, suggest a nearby park, or gamify outdoor 
activity through competition with peers. 
These just-in-time adaptive interventions leverage the continuous data 
stream from Section 4 to deliver behavioural support exactly when it is 
most needed \citep{wong2020}.

%% --- Figure 5 placeholder: Personalized intervention decision flow ---
%% Please insert: Decision flowchart showing the personalized intervention pathway:
%%   Input: Risk profile (from Sec 3) + Monitoring data (from Sec 4)
%%   → AI treatment recommender (atropine/OK/MFCL/digital, dose selection)
%%   → Intervention delivery
%%   → Monitoring response (closed loop back to Sec 4)
%%   Include decision node for when to escalate/de-escalate/switch treatment.
\begin{figure*}[htbp!]
\centering
\includegraphics[width=\textwidth]{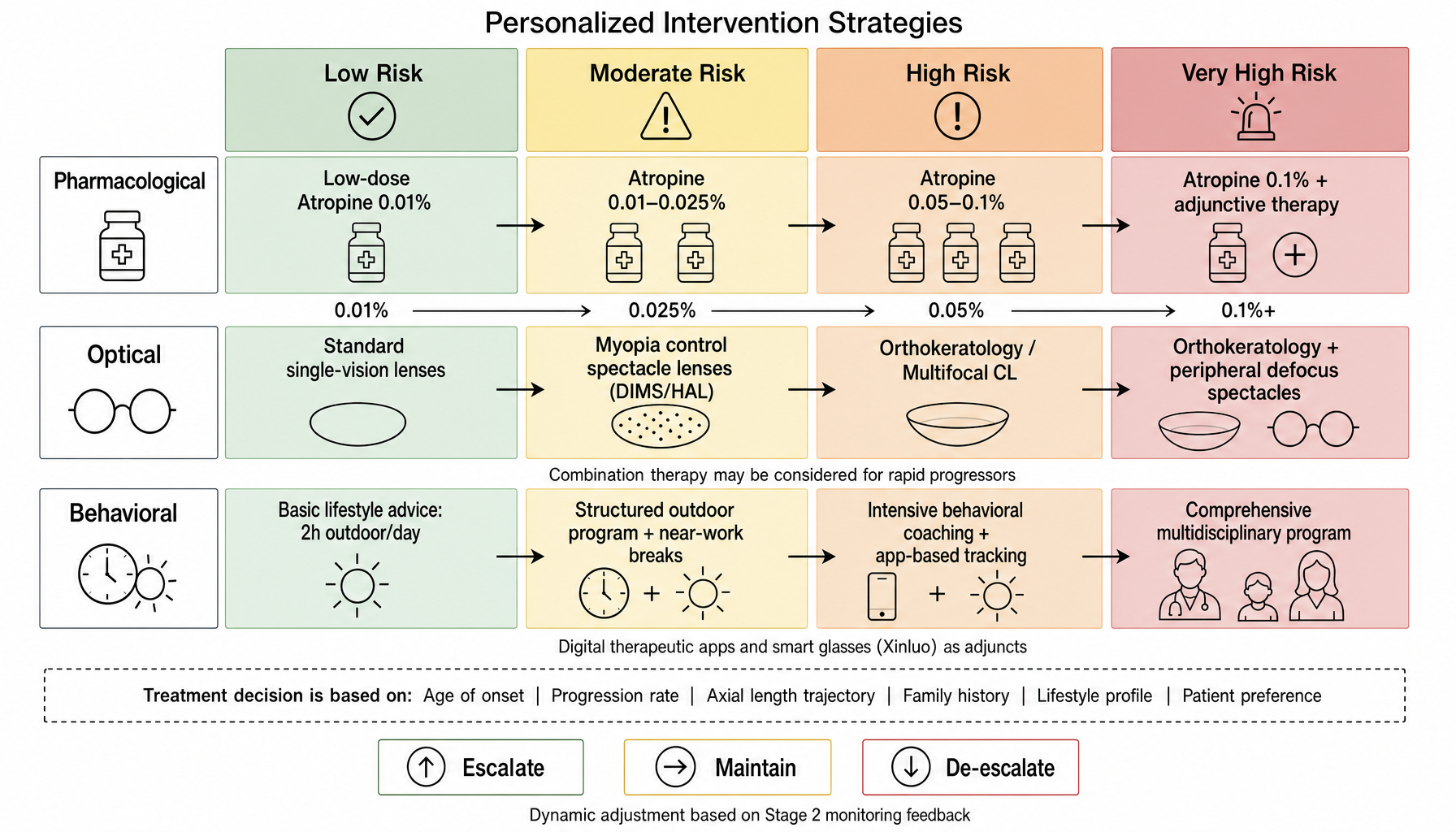}
\caption{%
  Personalised intervention framework. Risk profile and monitoring data
  inform treatment selection and dose titration. Intervention response is
  continuously assessed, creating a closed-loop system that escalates, maintains,
  or de-escalates therapy based on individual trajectories.}
\label{fig:intervention}
\end{figure*}

\subsection{Closing the Loop: Monitoring Feedback and Dynamic Adjustment}
\label{sec:5.4}

The defining feature of the 3.0 paradigm---the element that most clearly 
distinguishes it from 2.0---is the closed feedback loop from intervention 
(Stage 5) to monitoring (Stage 4). After an intervention is initiated or 
modified, the monitoring system continues to collect data on behavioural, 
environmental, and ocular parameters, evaluating whether the intended 
effect is being achieved. If the monitoring data indicate suboptimal 
response (e.g., axial elongation continues at $>$ 0.3 mm/year despite 
treatment), the system signals the need for intervention adjustment---
increasing atropine concentration, adding a second modality, or 
intensifying behavioural support.

This adaptive treatment paradigm is conceptually distinct from the 
standard ``treat and review'' model at fixed intervals. In 2.0, a child 
started on atropine 0.01\% might be reviewed at six months; if progression 
was rapid, the concentration would be increased at that visit. In 3.0, 
the monitoring system can detect rapid progression within weeks via 
continuous axial length measurement (where available) or behavioural 
surrogates (reduced outdoor time, increased near-work), triggering an 
earlier review. The system can also detect the opposite scenario---a child 
with excellent response at a lower dose---potentially allowing de-
escalation of therapy, reducing side effects and cost.

Implementing this closed-loop system in practice requires three 
technical components: (1) a quantitative metric of treatment response 
(most practically, axial length change rate normalised for age); (2) a 
decision algorithm that specifies when to escalate, maintain, or de-
escalate treatment based on this metric; and (3) a communication system 
that delivers actionable recommendations to clinicians and families 
without adding excessive alert burden. Early implementations of these 
components exist in clinical trial settings, but their deployment in 
real-world myopia care remains a key challenge for the field.

\subsection{Integration with Clinical Decision Support}
\label{sec:5.5}

For the 3.0 paradigm to translate from research into practice, its 
AI-driven recommendations must integrate seamlessly into clinical 
workflows. Clinical decision support (CDS) systems that embed risk 
predictions, monitoring summaries, and treatment recommendations into 
the electronic health record (EHR) can facilitate this integration. At the 
point of care, the CDS would present the clinician with: (1) the child's 
current risk profile, updated with the latest monitoring data; (2) the 
trajectory of key parameters (axial length, spherical equivalent, outdoor 
time) over time; (3) AI-generated treatment recommendations with 
evidence summaries and confidence estimates; and (4) suggested recall 
intervals based on risk trajectory.

Crucially, the CDS must support---rather than replace---clinical 
judgement. Shared decision-making, in which the AI provides evidence 
and the clinician and family make the final choice, is essential for 
maintaining trust and autonomy. The challenge of designing AI systems 
that are transparent (explaining why a recommendation was made), 
accountable (allowing clinician override), and equitable (avoiding bias 
across demographic groups) represents a cross-cutting priority that 
applies to all stages of the 3.0 pipeline \citep{pan2025}.

%% ====================================================================
%% 6. CHALLENGES AND FUTURE DIRECTIONS
%% ====================================================================
\section{Challenges and Future Directions}
\label{sec:challenges}

The success of the 3.0 paradigm depends on addressing cross-cutting 
challenges that span all three stages of the pipeline.

\subsection{Data Quality, Standardization, and Interoperability}
\label{sec:6.1}

Each stage of the 3.0 pipeline faces data-related challenges. In the risk 
stratification stage (Section 3), models require large, high-quality, 
well-labelled training datasets. Most published models have been trained 
on single-ethnicity cohorts, raising questions about generalisability to 
diverse populations. In the monitoring stage (Section 4), device 
heterogeneity---different actigraphy models, smartphone platforms, and 
measurement protocols---introduces noise that degrades the performance 
downstream models. In the intervention stage (Section 5), treatment 
response prediction requires standardised outcome data collected under 
consistent protocols across multiple treatment modalities.

Addressing these challenges requires adoption of FAIR (Findable, 
Accessible, Interoperable, Reusable) data principles across the myopia 
research community. The establishment of shared data schemas for risk 
factors, monitoring outputs, and treatment outcomes would enable 
cross-study analyses, meta-learning across datasets, and more robust 
model validation. Federated learning---where models are trained across 
multiple institutions without sharing raw data---offers a promising 
approach to building more generalisable models while respecting data 
privacy and institutional boundaries.

\subsection{Model Validation, Reproducibility, and Generalizability}
\label{sec:6.2}

The rapid proliferation of ML models for myopia prediction and 
intervention optimisation has outpaced their rigorous validation. Many 
published studies lack external validation in independent cohorts, and 
fewer still have been validated across different ethnic groups, 
socioeconomic contexts, or health systems. As a result, it remains unclear 
which models will generalise to the diverse populations that would be 
served by a global 3.0 framework.

Several reporting guidelines have been developed to address these concerns 
and should be adopted by the field. The TRIPOD-AI (Transparent Reporting 
of a multivariable prediction model for Individual Prognosis Or Diagnosis-
Artificial Intelligence) statement provides a framework for reporting 
prediction model studies. PROBAST (Prediction model Risk Of Bias 
ASsessment Tool) enables critical appraisal of prediction models. The 
CLAIM (Checklist for Artificial Intelligence in Medical Imaging) checklist 
covers reporting standards specific to AI in medical imaging. Future 
studies should adhere to these guidelines, pre-register analysis plans, 
and make code and (where possible) data publicly available to enable 
independent replication. Prospective validation studies, in which the model 
is tested in real-time clinical workflows, represent the gold standard and 
are urgently needed before clinical deployment.

\subsection{Ethical Considerations, Privacy, and Equity}
\label{sec:6.3}

The continuous data collection inherent to the 3.0 paradigm raises 
important ethical and privacy concerns. Monitoring children's behaviour, 
location, and screen use through wearable devices and smartphones 
generates a digital footprint that could be misused if not properly 
safeguarded. Clear frameworks for data ownership, parental consent, and 
youth assent are needed, particularly as children reach adolescence and 
their preferences regarding data sharing may evolve.

Algorithmic bias poses another critical challenge. If prediction models 
are trained predominantly on East Asian populations---as most current 
models are---their performance in European, African, or Latin American 
populations may be substantially worse. Similarly, monitoring technologies 
(wearables, smartphones) are less accessible to lower-income populations, 
creating a digital divide that could exacerbate rather than reduce myopia-
related health disparities. The 3.0 paradigm must be designed with equity 
as a core principle, ensuring that AI advances benefit all children 
regardless of geography, ethnicity, or socioeconomic status.

Regulatory frameworks for AI-based preventive health tools are still 
evolving. Most current myopia AI models would be classified as clinical 
decision support tools, but those that operate autonomously (e.g., 
automated alert generation based on monitoring data) may face more 
stringent regulatory requirements. Early engagement with regulatory 
agencies and adherence to emerging AI governance standards will be 
essential for responsible translation of the 3.0 framework into practice.

\subsection{Clinical Integration and Health System Readiness}
\label{sec:6.4}

Translating the 3.0 pipeline from a conceptual framework into clinical 
reality requires health system readiness across multiple dimensions. 
Integration with electronic health records is a prerequisite for 
operationalising risk prediction and monitoring alerts in clinical 
workflows. Clinician training and trust in AI recommendations must be 
addressed through education, transparent model explanations, and 
demonstration of clinical utility in real-world settings. Reimbursement 
models need to evolve to cover continuous monitoring services and AI-
augmented clinical decision support, which do not fit easily into 
traditional fee-for-service payment structures.

Economic evaluations---including cost-effectiveness analyses of the 
entire 3.0 pipeline, not just individual components---are needed to 
justify investment in the required infrastructure. Early evidence 
suggests that AI-enhanced screening and monitoring can reduce overall 
costs by preventing progression to high myopia and its complications 
\citep{jin2025}, but comprehensive modelling across different health 
system contexts is lacking. Policy recommendations should address 
these gaps and provide a roadmap for phased implementation, starting 
with high-risk populations in well-resourced settings and progressively 
expanding as evidence accumulates and costs decrease.

\subsection{Future Directions}
\label{sec:6.5}

Looking ahead, several emerging technologies and approaches hold promise 
for advancing the 3.0 paradigm. First, multimodal foundation models---
trained across imaging, text, and sensor data---could unify the three 
stages of the pipeline within a single architecture, enabling more 
seamless integration and potentially uncovering cross-stage patterns that 
today's stage-specific models cannot capture. Second, digital twins---
computational models that represent an individual's eyes and visual system 
and simulate their response to different interventions---could enable 
``in silico'' treatment optimisation before any clinical intervention is 
initiated. Third, causal machine learning methods could move beyond 
associational prediction to identify causal drivers of progression for 
each individual, enabling more precisely targeted interventions 
\citep{pan2025}. Fourth, large language models could power patient-facing 
chatbots for education, adherence support, and triage, extending the reach 
of specialised myopia care beyond tertiary centres. Finally, edge AI---
processing data on wearable devices rather than in the cloud---could 
enable real-time, privacy-preserving monitoring without the need for 
continuous internet connectivity, addressing both privacy concerns and the digital divide.

%% --- Figure 6 placeholder: Challenges and future directions overview ---
%% Please insert: A visual summary of key challenges and future directions.
\begin{figure*}[htbp!]
\centering
\includegraphics[width=\textwidth]{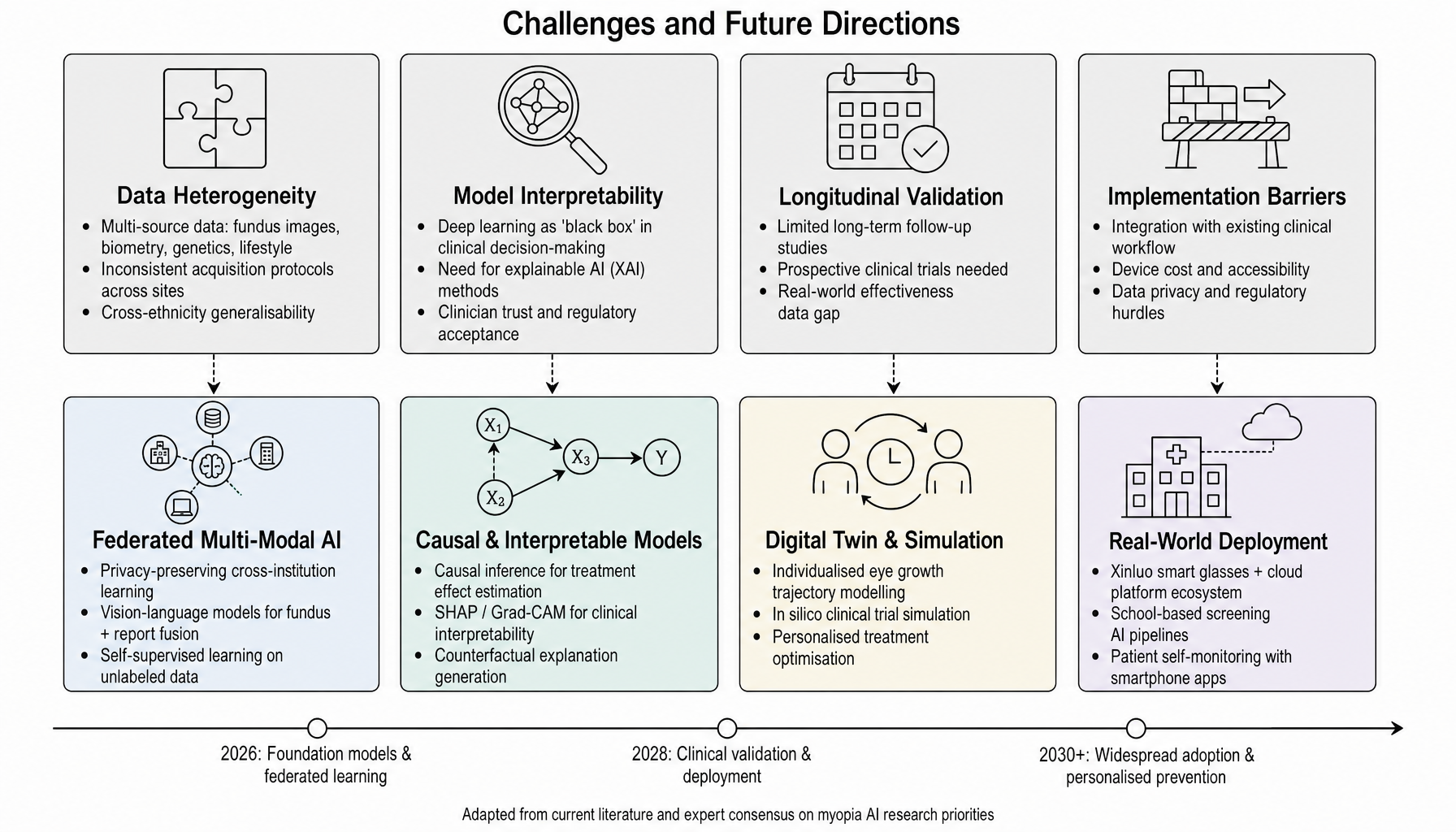}
\caption{%
  Challenges and future directions. Key challenge domains (data quality,
  model validation, ethics/equity, clinical integration, digital divide) are
  mapped against enabling technologies (foundation models, digital twins,
  causal ML, LLMs, edge AI).}
\label{fig:challenges}
\end{figure*}

%% ====================================================================
%% 7. CONCLUSION
%% ====================================================================
\section{Conclusion}
\label{sec:conclusion}

Myopia prevention and control stand at a pivotal juncture. The 
conventional approaches of population screening (1.0) and evidence-based 
risk management (2.0) have provided important foundations but are 
insufficient for the scale and complexity of the global myopia challenge. 
The 3.0 paradigm---an AI-driven, closed-loop pipeline linking risk 
stratification, proactive monitoring, and personalised intervention---
offers a coherent framework for the next generation of myopia prevention 
and control.

The evidence reviewed here demonstrates that the building blocks of this 
framework are already in place. AI prediction models can identify at-risk 
children with AUCs exceeding 0.85 across diverse clinical settings 
(Section 3). Digital sensing technologies can passively and continuously 
track behavioural risk factors with sufficient accuracy for clinical use 
(Section 4). Data-driven approaches can predict individual treatment 
response and support dynamic intervention adjustment (Section 5). What is 
needed now is the integration and operationalisation of these components 
into unified, scalable systems that operate across the full risk-to-
intervention pipeline, connected by the closed-loop feedback mechanism 
that is the defining innovation of the 3.0 paradigm.

Realising this vision will require concerted, interdisciplinary effort. 
Researchers must develop and validate models across diverse populations, 
adhering to rigorous reporting standards and pre-registered analysis 
plans. Clinicians must champion the clinical workflows and training 
programmes that will embed AI-enhanced care into routine practice. 
Engineers must build the scalable, interoperable platforms that connect 
screening, monitoring, and intervention data across healthcare settings. 
Policymakers must create the regulatory and reimbursement frameworks that 
incentivise innovation while protecting privacy and promoting equity. 
Educators and families must engage as active partners in the monitoring 
and behavioural modification process.

The trajectory is clear. Just as the 20th century saw the transition from 
treating established disease to preventing its development, the 21st 
century is witnessing the transition from uniform prevention to precision 
prevention. Myopia Prevention and Control 3.0---proactive, data-driven, 
and continuously adaptive---represents not merely an incremental advance 
but a fundamental re-imagination of what myopia care can achieve. The 
question is no longer whether this transformation is possible, but how 
rapidly and equitably we can bring it to the children who need it most.

%% ====================================================================
%% REFERENCES
%% ====================================================================
\raggedbottom
\bibliographystyle{elsarticle-num}
\bibliography{refs}

\end{document}